\documentclass{article}

\usepackage{microtype}
\usepackage{graphicx}
\usepackage{subfigure}
\usepackage{booktabs} % for professional tables

\usepackage{hyperref}

\usepackage[accepted]{icml2025}

\makeatletter
\renewcommand{\Notice@String}{}
\makeatother
\hypersetup{
  pdfsubject={arXiv preprint}
}

\usepackage{amsmath}
\usepackage{amssymb}
\usepackage{mathtools}
\usepackage{amsthm}

\usepackage[capitalize,noabbrev]{cleveref}

\theoremstyle{plain}

\theoremstyle{definition}

\theoremstyle{remark}

\usepackage[textsize=tiny]{todonotes}

\icmltitlerunning{Physics2Weight}

\begin{document}

\twocolumn[
\icmltitle{Physics2Weight: Direct Neural Network Architecture and Weight Generation from Physical Laws Using Large Language Models}

\begin{icmlauthorlist}
\icmlauthor{Shogo Sato}{ntt}
\icmlauthor{Tomoyasu Shimada}{ntt}
\icmlauthor{Kazuhiko Murasaki}{ntt}
\icmlauthor{Takashi Sano}{ntt}
\end{icmlauthorlist}

\icmlaffiliation{ntt}{NTT, Inc., Japan}
\icmlcorrespondingauthor{Shogo Sato}{shg.sato@ntt.com}
\icmlkeywords{Physics-Informed Machine Learning, Large Language Models, Weight-Space Programming, Partial Differential Equations}

\vskip 0.3in
]

\printAffiliationsAndNotice{\icmlEqualContribution}

\begin{abstract}
% P1. 背景：NNの重みは通常、勾配最適化によって獲得されるが、WSPはLLMによる直接的な重み生成を可能にする。
Recent advances in large language models (LLMs) have enabled weight-space programming (WSP), which directly generates neural network weights without gradient-based optimization.
% P2. 課題：WSPを物理問題に適用する際には、解の近似精度と物理的整合性の両立が課題となる。
However, extending WSP to physical problems remains challenging, particularly in achieving both solution accuracy and physical consistency.
% P3. 提案：PtWは支配方程式と初期・境界条件からNNの重みとアーキテクチャを直接生成する。
In this paper, we propose Physics2Weight (PtW), a framework that extends WSP to solving partial differential equations by directly generating neural network weights and editing architectures based on governing equations and initial and boundary conditions.
% P4. 手法：物理制約プロンプトとアーキテクチャ編集を導入し、既存WSPの反復的重み修正と組み合わせる。
PtW introduces physics-constrained prompting and physics-guided architecture editing, combined with iterative weight refinement from WSP, to improve solution accuracy and physical consistency without gradient-based optimization.
% P5. 結果：4種類のPDEでPINNより高い解の近似精度を達成する一方、非線形PDEでは物理整合性に課題が残る。
Experiments on four PDEs, namely the heat, advection, Burgers', and Allen--Cahn equations, demonstrate lower solution errors than conventional physics-informed neural networks, while revealing limitations in satisfying the governing equations for nonlinear PDEs.
\end{abstract}

\begin{figure}
\centering
\includegraphics[width=\columnwidth]{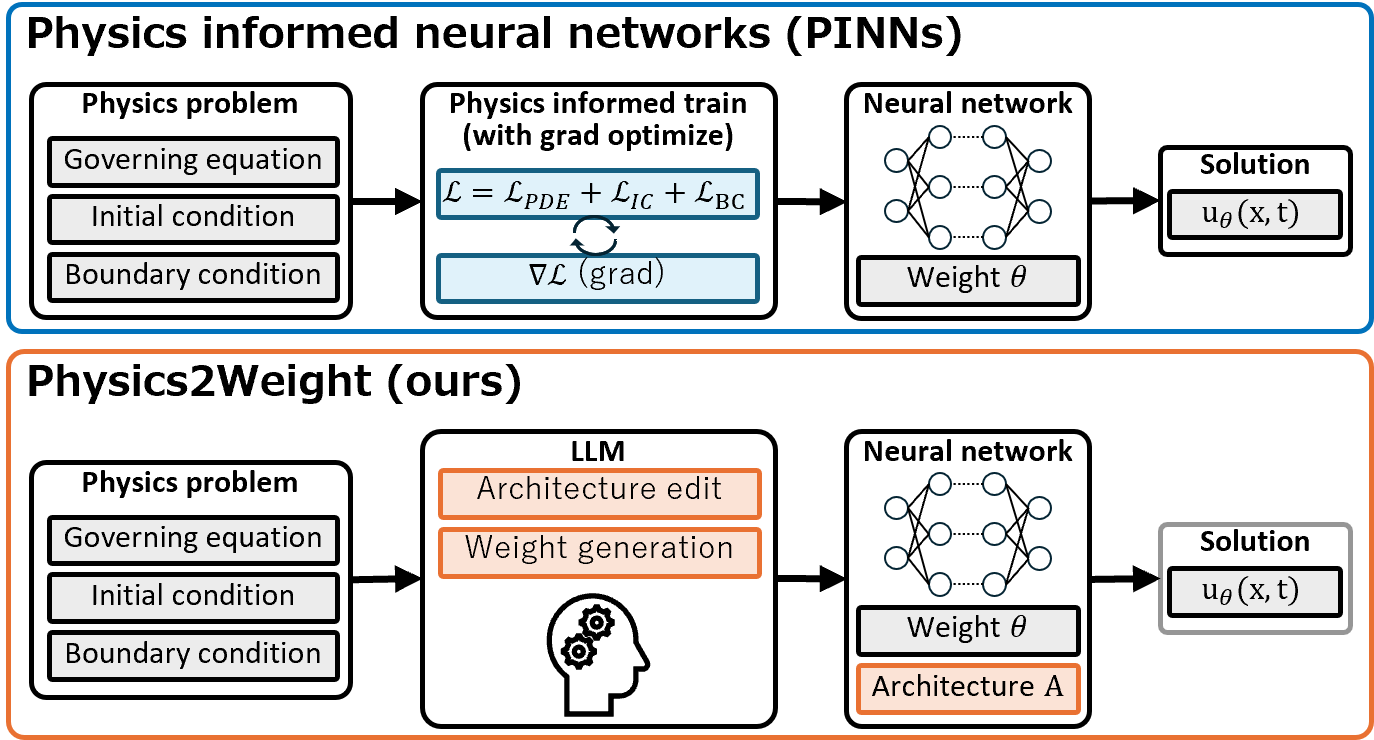}
\vspace{-7mm}
\caption{Conceptual comparison between (a) conventional physics-informed neural networks (PINNs) and (b) the proposed Physics2Weight (PtW). PINNs optimize neural network weights by minimizing physics-informed loss functions, whereas PtW edits architectures and directly generates weights from governing equations and initial and boundary conditions using an LLM, without gradient-based optimization.}
\label{fig:overview}
\end{figure}

\section{Introduction}\label{sec:intro}

% P1. 背景：NNは幅広い分野で高い性能を示しているが、その重みの獲得には一般に学習データや勾配ベースの最適化が必要となる。
Neural networks (NNs) have demonstrated strong performance across a wide range of tasks, including computer vision, natural language processing, and scientific computing~\cite{he2016deep, vaswani2017attention, raissi2019physics}.
However, obtaining their weights typically relies on gradient-based optimization, often using large amounts of training data or task-specific loss functions.

% P2. 新しい潮流：WSPはLLMによってNNの重みを直接生成し、従来の勾配最適化を介さないNN構築を可能にする。
Recent advances in large language models (LLMs)~\cite{vaswani2017attention, brown2020language} have opened up the possibility of weight-space programming (WSP)~\cite{takehi2026weightspace}, in which LLMs directly generate NN weights without gradient-based optimization.
This approach offers an alternative paradigm for constructing NNs by generating their parameters rather than learning them through conventional training.

% P3. 物理問題への展開と課題：PINNは物理法則を損失関数に組み込んで重みを最適化するが、WSPによる直接的な重み生成で解の精度と物理整合性を両立できるかは明らかでない。
Solving partial differential equations (PDEs) is a potential application of WSP, as physical problems are often specified by governing equations and initial and boundary conditions.
Physics-informed neural networks (PINNs)~\cite{raissi2019physics} incorporate these conditions into loss functions and optimize NN weights through gradient-based training.
In contrast, whether WSP can directly generate NN weights that achieve both accurate solution approximation and physical consistency remains unclear.

% P4. 提案手法：PtWは物理制約プロンプトとアーキテクチャ編集を導入し、既存WSPの反復的重み修正と組み合わせてPDEを解く。
In this paper, we propose Physics2Weight (PtW), a framework that extends WSP to solving PDEs by directly generating NN weights using LLMs (Fig.~\ref{fig:overview}).
PtW introduces two physics-guided strategies: (1) physics-constrained prompting, which explicitly specifies additional physical constraints to guide weight generation, and (2) physics-guided architecture editing, which incorporates mathematical structures of governing equations into NN architectures.
These strategies are combined with an iterative weight refinement procedure based on WSP~\cite{takehi2026weightspace} to improve the generated solutions without gradient-based optimization.

% P5. 実験：4種類のPDEでPINNおよびWSPと比較し、解の精度と物理整合性を評価する。
We evaluate PtW on four PDEs: heat, advection, Burgers', and Allen--Cahn equations.
Through comparisons with PINNs and WSP and ablation studies, we quantitatively assess solution accuracy and physical consistency.
The results show that PtW achieves lower solution errors than PINNs and WSP across all four PDEs, although improvements in solution accuracy do not always correspond to lower PDE residuals.

% P6. 貢献：PtWの提案、物理制約プロンプトとアーキテクチャ編集の検討、および4種類のPDEにおける体系的な評価を主な貢献とする。
Our contributions are summarized as follows:
\begin{itemize}
\item We introduce PtW, a framework for solving PDEs through direct NN weight generation using LLMs, without gradient-based optimization.
\item We investigate physics-constrained prompting and physics-guided architecture editing, combined with iterative weight refinement, to improve solution accuracy and physical consistency.
\item We systematically evaluate PtW across four PDEs, demonstrating improved solution accuracy over PINNs and WSP while identifying limitations in physical consistency.
\end{itemize}

\section{Related Work}\label{sec:related}

% P1. Neural Network Weight Generation：学習ベースの重み生成手法とWSPを整理し、PtWによる物理制約とアーキテクチャ編集の導入を位置づける。
\noindent\textbf{Neural Network Weight Generation.}
NN weight generation aims to produce the parameters of target NNs without directly optimizing each target network.
Hypernetworks~\cite{ha2017hypernetworks} generate target NN weights using a separately trained network.
Weight-space learning~\cite{navon2023equivariant} further investigates the structure and representations of NN parameters, while recent generative approaches~\cite{wang2024neural, soro2025diffusion, kim2026nnit} explore the generation of NN weights from learned parameter distributions.
Text-conditioned approaches extend this direction by generating weights from natural language descriptions~\cite{tian2025text2weight}.
More recently, WSP~\cite{takehi2026weightspace} has reported that general-purpose LLMs can directly generate and refine NN weights without training a dedicated weight generator.
Building on WSP, PtW introduces explicit physical constraints and physics-guided architecture editing to construct NN representations for solving PDEs.\smallskip

% P2. Neural Architecture Search and Editing：従来のNASとLLMによるアーキテクチャ探索を整理し、PtWの物理法則に基づく計算構造の編集との差異を示す。
\noindent\textbf{Neural Architecture Search and Editing.}
Neural architecture search (NAS) aims to automatically identify effective NN architectures for a given task.
Early NAS methods employ reinforcement learning~\cite{zoph2016neural} or gradient-based optimization~\cite{liu2018darts} to explore architecture search spaces.
Recent LLM-based approaches, such as GENIUS~\cite{zheng2023can} and LLMatic~\cite{nasir2024llmatic}, generate and iteratively refine NN architectures using performance feedback.
RZ-NAS~\cite{ji2025rz} further incorporates LLM-based reflection to guide architecture search.
Unlike general-purpose NAS, PtW uses the mathematical structure of governing equations to guide architecture editing, explicitly incorporating physically motivated operations into NN computations.\smallskip

% P3. Physics-Informed Neural Networks：PINNと物理制約を厳密に組み込む既存手法を整理し、PtWによるLLMベースの構造編集と重み生成の統合との差異を示す。
\noindent\textbf{Physics-Informed Neural Networks.}
PINNs incorporate governing equations and initial and boundary conditions into loss functions to approximate PDE solutions without requiring extensive labeled solution data~\cite{raissi2019physics}.
Earlier work enforced boundary conditions exactly through trial functions~\cite{lagaris1998artificial}.
Subsequent studies have investigated adaptive loss balancing~\cite{wang2021understanding}, domain
decomposition~\cite{jagtap2020extended}, and exact imposition of boundary conditions with distance
functions~\cite{sukumar2022exact} to improve training and constraint satisfaction.
In particular, hard-constrained approaches incorporate analytical structures into NN representations to satisfy prescribed physical conditions.
Unlike these methods, PtW uses LLMs to edit NN architectures based on governing equations and directly generate and refine their weights without gradient-based optimization.

\begin{figure*}[t]
\centering
\includegraphics[width=0.8\textwidth]{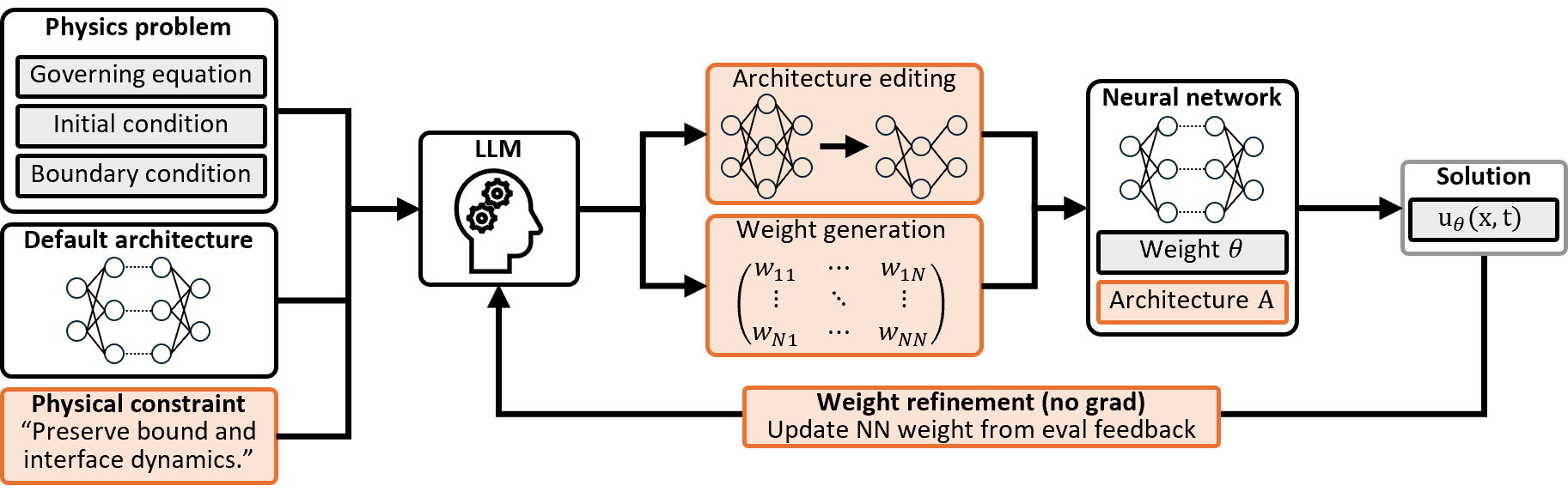}
\vspace{-4mm}
\caption{Overview of the proposed PtW framework. Given governing equations, initial and boundary conditions, and an initial NN architecture, an LLM edits the architecture and directly generates NN weights without gradient-based optimization. Physics-constrained prompting specifies additional physical constraints, while iterative weight refinement, adapted from WSP, updates the generated weights based on external evaluation feedback.}
\label{fig:framework}
\end{figure*}

\section{Preliminaries}\label{sec:preliminaries}

\subsection{Problem Formulation}

% P1. 物理問題の定義：支配方程式、初期条件、境界条件によって定義されるPDEの初期境界値問題を数学的に定式化する。
We consider an initial-boundary value problem governed by a PDE:
\begin{equation}
\begin{aligned}
\mathcal{F}[u](\mathbf{x},t) &= 0,
&& (\mathbf{x},t)\in\Omega\times(0,T],\\
\mathcal{I}[u](\mathbf{x},0) &= 0,
&& \mathbf{x}\in\Omega,\\
\mathcal{B}[u](\mathbf{x},t) &= 0,
&& (\mathbf{x},t)\in\partial\Omega\times(0,T],
\end{aligned}
\label{eq:pde}
\end{equation}
where $u(\mathbf{x},t)$ denotes the unknown solution, $\mathbf{x}\in\Omega$ is the spatial coordinate, $t\in[0,T]$ is time, and $\mathcal{F}$, $\mathcal{I}$, and $\mathcal{B}$ represent the governing equation, initial condition, and boundary condition operators, respectively.

% P2. NNによる解の表現：NNのアーキテクチャAと重みthetaを定義し、PDEの解をNNで近似する問題として定式化する。
We approximate the solution using an NN $u_{\boldsymbol{\theta},\mathcal{A}}(\mathbf{x},t)$, parameterized by weights $\boldsymbol{\theta}$ and architecture $\mathcal{A}$:
\begin{equation}
u(\mathbf{x},t)\approx u_{\boldsymbol{\theta},\mathcal{A}}(\mathbf{x},t).
\label{eq:nn_solution}
\end{equation}
The objective is to construct an NN representation that accurately approximates the solution while satisfying the governing equation and initial and boundary conditions.

\subsection{Gradient-based Weight Optimization}

% P1. 物理損失関数：NNに時空間座標を入力して近似解を得て、支配方程式の残差、初期条件、境界条件に基づくPINNの損失関数を定義する。
Given spatiotemporal coordinates $(\mathbf{x},t)$, PINNs predict the approximate solution $u_{\boldsymbol{\theta},\mathcal{A}}(\mathbf{x},t)$ and minimize a physics-informed loss function consisting of governing equation residuals and initial and boundary condition errors:
\begin{equation}
\mathcal{L}(\boldsymbol{\theta};\mathcal{A})
=\lambda_{\mathrm{PDE}}\mathcal{L}_{\mathrm{PDE}}
+\lambda_{\mathrm{IC}}\mathcal{L}_{\mathrm{IC}}
+\lambda_{\mathrm{BC}}\mathcal{L}_{\mathrm{BC}},
\label{eq:pinn_loss}
\end{equation}
where $\mathcal{L}_{\mathrm{PDE}}$, $\mathcal{L}_{\mathrm{IC}}$, and $\mathcal{L}_{\mathrm{BC}}$ measure violations of the governing equation, initial condition, and boundary condition, respectively, and $\lambda_{\mathrm{PDE}}$, $\lambda_{\mathrm{IC}}$, and $\lambda_{\mathrm{BC}}$ are their weighting coefficients.

% P2. 重み最適化：固定アーキテクチャの下で損失関数を勾配最適化し、NNの重みを獲得する従来手法を説明する。
Given a fixed architecture $\mathcal{A}$, PINNs obtain NN weights by minimizing the physics-informed loss function through gradient-based optimization:
\begin{equation}
\boldsymbol{\theta}^{*}
=\arg\min_{\boldsymbol{\theta}}\mathcal{L}(\boldsymbol{\theta};\mathcal{A}).
\label{eq:pinn_optimization}
\end{equation}
This process iteratively updates the weights using gradients of the loss function with respect to $\boldsymbol{\theta}$.

\subsection{Weight-Space Programming}

% P1. WSPの概要：LLMが勾配最適化なしにNNの重みを直接生成する。
WSP enables pretrained LLMs to directly generate NN weights from a specification of the target network without gradient-based optimization~\cite{takehi2026weightspace}.
Given an NN architecture $\mathcal{A}$ and a task specification $\mathcal{S}$, the generated weights are expressed as
\begin{equation}
\boldsymbol{\theta}^{(0)}=\mathrm{LLM}(\mathcal{S},\mathcal{A}),
\end{equation}
where $\mathcal{S}$ describes the target task and $\boldsymbol{\theta}^{(0)}$ denotes the initially generated weights.

% P2. WSPの反復修正：LLM自身がforward passを実行し、その結果を用いて重みを修正する。
WSP also supports iterative weight refinement~\cite{takehi2026weightspace}.
The LLM evaluates the generated network by executing forward passes and uses the resulting feedback to modify its weights.
At iteration $k$, this process can be expressed as
\begin{equation}
\boldsymbol{\theta}^{(k+1)}
=\mathrm{LLM}(\mathcal{S},\mathcal{A},
\boldsymbol{\theta}^{(k)},\mathcal{E}^{(k)}),
\end{equation}
where $\mathcal{E}^{(k)}$ denotes the evaluation feedback obtained by the LLM.
This process iteratively refines NN weights without computing gradients of the evaluation objective.
In contrast, PtW adapts this refinement procedure to PDE solving by providing physics-based feedback from an external evaluator, without granting the LLM tool access.

\begin{figure*}[t]
\centering
\includegraphics[width=1.0\textwidth]{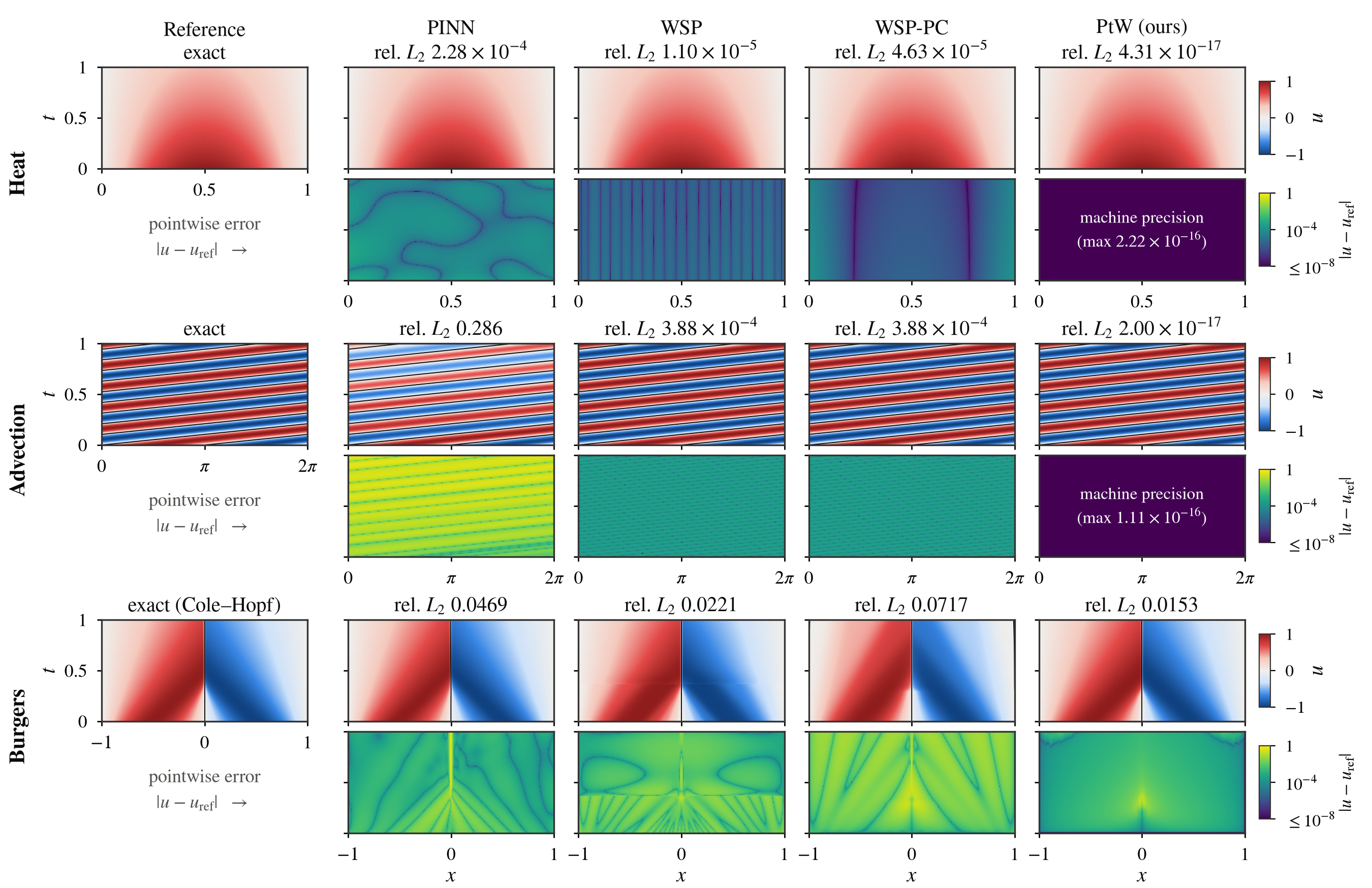}
\vspace{-8mm}
\caption{Comparison of spatiotemporal solutions and pointwise absolute errors for the heat, advection, and Burgers' equations. Reference solutions are compared with predictions from PINN, WSP, WSP-PC, and PtW. For each equation, the top row shows the predicted solutions, while the bottom row shows the pointwise absolute errors on a logarithmic color scale shared across all panels, with errors below $10^{-8}$ displayed using the lowest color. Black lines indicate $u=0$, and each method is represented by the run with the median relative $L_2$ error among three runs.}
\label{fig:solutions}
\end{figure*}

\section{Physics2Weight}\label{sec:method}

\subsection{Framework Overview}

% P1. 全体構成：支配方程式、初期・境界条件、物理制約、初期NNアーキテクチャを入力として、PC・AE・WRを組み合わせてNNを構築する。
As illustrated in Fig.~\ref{fig:framework}, PtW constructs an NN for solving PDEs using a pretrained LLM without gradient-based optimization.
Given the governing equation $\mathcal{F}$, initial condition $\mathcal{I}$, boundary condition $\mathcal{B}$, physical constraints $\mathcal{C}$, and an initial NN architecture $\mathcal{A}_0$, PtW generates an architecture $\widehat{\mathcal{A}}$ and weights $\widehat{\boldsymbol{\theta}}$:
\begin{equation}
(\widehat{\mathcal{A}},\widehat{\boldsymbol{\theta}})
=\mathrm{PtW}(\mathcal{F},\mathcal{I},\mathcal{B},\mathcal{C},\mathcal{A}_0).
\label{eq:ptw_framework}
\end{equation}
PtW combines three components: physics-constrained prompting (PC), physics-guided architecture editing (AE), and iterative weight refinement (WR).
PC explicitly specifies physical constraints, while AE incorporates mathematical structures of governing equations into NN architectures.
WR, adapted from WSP~\cite{takehi2026weightspace}, iteratively refines the generated weights based on external evaluation feedback.
The complete PtW framework combines all three components, while their individual effects are investigated through ablation studies.

\subsection{Physics-Constrained Prompting}\label{sec:pc}

% P1. 物理制約プロンプト：通常のWSPに追加の物理制約を与え、解の精度と物理的整合性の両方を評価する。
In standard WSP, the task specification $\mathcal{S}$ includes the governing equation, computational domain, initial and boundary conditions, NN architecture, and evaluation criteria.
Physics-constrained prompting (PC) additionally specifies physical constraints that the exact solution is known to satisfy.
These constraints include initial and boundary condition tolerances, maximum principles, conservation or dissipation properties, and problem-specific physical properties.
We construct a physics-constrained prompt $\mathcal{S}_{\mathrm{PC}}$ by augmenting the original task specification with a set of constraints $\mathcal{C}$:
\begin{equation}
\boldsymbol{\theta}^{(0)}
=\mathrm{LLM}(\mathcal{S}_{\mathrm{PC}},\mathcal{A}_0),
\quad
\mathcal{S}_{\mathrm{PC}}=(\mathcal{S},\mathcal{C}).
\label{eq:pc_generation}
\end{equation}
The prompt states that a network violating any constraint is regarded as physically incorrect even if its solution error is small, and that solution accuracy and physical fidelity are evaluated together.
During iterative weight refinement, constraint violations are evaluated without reference solutions and incorporated into the feedback and model selection.
The physical constraints used for each PDE are detailed in Appendix~\ref{app:constraints}.

\subsection{Physics-Guided Architecture Editing}\label{sec:ae}

% P1. アーキテクチャ編集：支配方程式の数学的構造に基づいて初期NNアーキテクチャを編集し、物理的な演算をforward computationに組み込む。
Physics-guided architecture editing (AE) allows the LLM to modify the initial NN architecture based on the mathematical structure of the governing equation and its initial and boundary conditions.
Rather than restricting weight generation to a predefined architecture $\mathcal{A}_0$, the LLM incorporates analytical operations and physically motivated transformations into the NN representation.
For example, exponential decay in the heat equation and characteristic coordinate transformations in the advection equation can be embedded into the forward computation.
The edited architecture is expressed as
\begin{equation}
\widehat{\mathcal{A}}
=\mathrm{LLM}(\mathcal{S}_{\mathrm{PC}},\mathcal{A}_0;\mathcal{R}),
\label{eq:ae_architecture}
\end{equation}
where $\mathcal{R}$ denotes architecture editing rules that restrict the available operations and parameter count while preserving trainable NN components.

% P2. 重み生成：編集後のアーキテクチャに対応する重みをLLMが直接生成し、実装上の制約を満たすことを確認する。
The edited architecture consists of mathematical expressions and MLP blocks, allowing physical knowledge to be incorporated into the network structure.
The generated architecture is validated against $\mathcal{R}$, and any violations are returned to the LLM for correction.
Given the edited architecture $\widehat{\mathcal{A}}$, the LLM directly generates its weights:
\begin{equation}
\boldsymbol{\theta}^{(0)}
=\mathrm{LLM}(\mathcal{S}_{\mathrm{PC}},\widehat{\mathcal{A}}).
\label{eq:ae_weights}
\end{equation}
This enables the NN to represent physical solutions through both its architecture and weights without gradient-based optimization.
The architecture editing rules and supported operations are described in Appendix~\ref{app:ae}.

\subsection{Iterative Weight Refinement}\label{sec:wr}

% P1. 反復的重み修正：WSPのWRを利用し、外部評価による物理損失と制約違反のフィードバックに基づいて重みを直接修正する。
WR follows the iterative refinement loop of WSP, but differs in its evaluation procedure: the LLM has no tool access and receives physics-based feedback from an external evaluator.
Given the generated weights $\boldsymbol{\theta}^{(k)}$, the NN is evaluated using the physics-informed loss in Eq.~\eqref{eq:pinn_loss}, including PDE residuals and initial and boundary condition errors.
The evaluation also provides spatial and temporal residual diagnostics and, when PC is enabled, the results of physical constraint checks.
These evaluations are performed externally without providing reference solutions or solution errors to the LLM.
At iteration $k$, the LLM updates the weights as
\begin{equation}
\boldsymbol{\theta}^{(k+1)}
=\mathrm{LLM}(\mathcal{S}_{\mathrm{PC}},\widehat{\mathcal{A}},
\boldsymbol{\theta}^{(k)},\mathcal{E}^{(k)}),
\label{eq:ptw_refinement}
\end{equation}
where $\mathcal{E}^{(k)}$ denotes the evaluation feedback.
The architecture remains fixed during refinement, and the weights are modified without gradient-based optimization.

% P2. モデル選択：物理制約違反数を優先し、同数の場合には物理損失に基づいて最終モデルを選択する。
The final weights are selected from the generated candidates based on physical constraint satisfaction and the physics-informed loss.
Specifically, PtW prioritizes the number of violated constraints $V$ and uses the physics-informed loss $\mathcal{L}$ to distinguish candidates with the same number of violations:
\begin{equation}
\widehat{\boldsymbol{\theta}}
=\boldsymbol{\theta}^{(k^*)},
\quad
k^*=\operatorname*{arg\,min}_{0\leq k\leq K}
\left(
V(\boldsymbol{\theta}^{(k)}),
\mathcal{L}(\boldsymbol{\theta}^{(k)})
\right),
\label{eq:wr_selection}
\end{equation}
where the minimization follows lexicographic ordering.
Without PC, the final weights are selected solely by minimizing $\mathcal{L}$.
This selection strategy prioritizes physical constraint satisfaction rather than solution accuracy alone.

\section{Experiments}\label{sec:experiments}

\subsection{Experimental Setup}

% P1. 評価タスク：線形・非線形の4種類のPDEを用いてPtWを評価し、詳細な初期・境界条件はAppendixに示す。
We evaluate PtW on four time-dependent PDEs: heat, advection, Burgers', and Allen--Cahn equations, covering diffusion, transport, nonlinear convection-diffusion, and reaction-diffusion.
Their governing equations are given by
\begin{equation}
\begin{aligned}
\text{Heat:}\quad &u_t=\alpha u_{xx},\\
\text{Advection:}\quad &u_t+\beta u_x=0,\\
\text{Burgers':}\quad &u_t+u u_x=\nu u_{xx},\\
\text{Allen--Cahn:}\quad &u_t=d\,u_{xx}+\kappa\,(u-u^3),
\end{aligned}
\label{eq:benchmark_pdes}
\end{equation}
with $\alpha=0.1$, $\beta=30$, $\nu=0.01/\pi$, $d=10^{-3}$, and $\kappa=5$.
We use $K=5$ refinement rounds, so the final weights are selected from $K+1=6$ candidates,
including the initially generated weights.
The corresponding physical parameters, computational domains, and initial and boundary conditions are provided in Appendix~\ref{app:pde_settings}.

% P2. 比較手法：PINN、WSPをベースラインとし、PC・AE・WRの各構成要素を組み合わせたアブレーションによりPtWを評価する。
We compare PtW with PINN and WSP as baseline methods.
PINN obtains NN weights through gradient-based optimization, whereas WSP directly generates weights using an LLM.
To investigate the effects of individual components, we evaluate PINN-WR, WSP-PC, WSP-WR, WSP-PC-WR, and WSP-PC-AE.
The complete PtW framework combines PC, AE, and WR.
Here, AE is evaluated together with PC, while WR can be applied independently of PC and AE.

% P3. 評価指標：相対L2誤差によって解の近似精度を評価し、PDE残差と初期・境界条件の誤差によって物理整合性を評価する。
We evaluate solution accuracy using the relative $L_2$ error between the predicted and reference solutions:
\begin{equation}
\mathcal{E}_{\mathrm{rel}}
=\frac{\|u_{\boldsymbol{\theta},\mathcal{A}}-u_{\mathrm{ref}}\|_2}
{\|u_{\mathrm{ref}}\|_2},
\label{eq:relative_l2}
\end{equation}
where $u_{\mathrm{ref}}$ denotes the reference solution evaluated at the same points.
Physical consistency is evaluated using the mean squared PDE residual and initial and boundary condition errors, denoted by $\mathcal{E}_{\mathrm{PDE}}$, $\mathcal{E}_{\mathrm{IC}}$, and $\mathcal{E}_{\mathrm{BC}}$, respectively.
Reference solutions are used only for independent evaluation and are not provided to the LLM.

% P4. 実装条件：使用するLLM、初期NNアーキテクチャ、評価条件、試行回数を説明する。
We use Claude Opus 5.5 for all LLM-based methods.
The initial architecture is a fully connected NN with two hidden layers of 20 $\tanh$ units, which is also used for the PINN baseline.
All LLM-based methods generate weights without gradient-based optimization.
Each experiment is repeated three times, and the median performance is reported.
Additional implementation details are provided in Appendix~\ref{app:implementation}.

\begin{figure*}[t]
\centering
\includegraphics[width=1.0\textwidth]{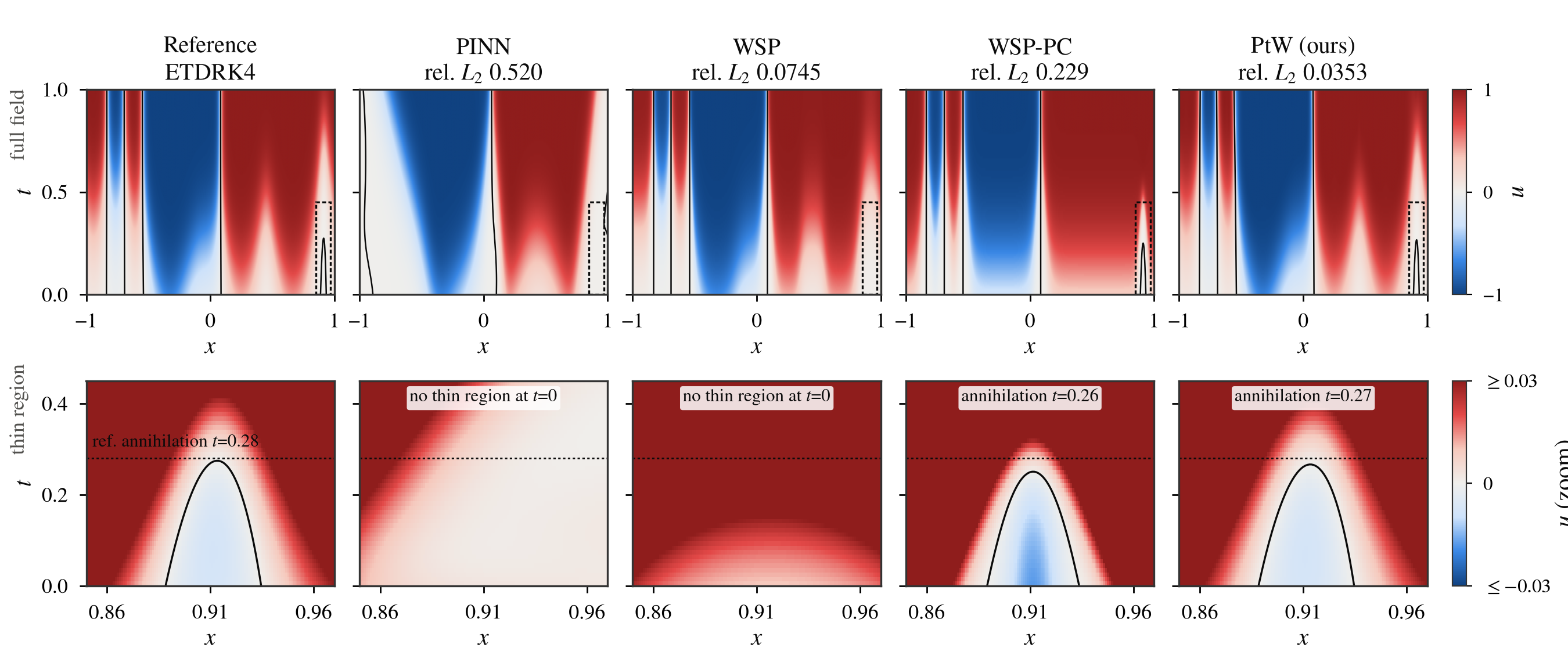}
\vspace{-8mm}
\caption{Comparison of spatiotemporal solutions for the Allen--Cahn equation. The bottom panels provide enlarged views of the interface dynamics, highlighting differences in interface evolution and annihilation timing between the reference solution and the evaluated methods. 
Each method is represented by the run with the median relative $L_2$ error among three runs. The dotted line indicates the reference annihilation time ($t=0.28$), and the enlarged panels use a color range of $[-0.03, 0.03]$.}
\label{fig:allen_cahn}
\end{figure*}

\begin{table}
\caption{Quantitative results on the heat equation. Values are medians over three runs.}
\label{tab:heat}
\centering
\scriptsize
\setlength{\tabcolsep}{3pt}
\begin{tabular}{lcccc}
\toprule
Method & Rel. L2 $\downarrow$ & PDE Res. $\downarrow$ & IC $\downarrow$ & BC $\downarrow$ \\
\midrule
PINN & $2.28\times10^{-4}$ & $6.45\times10^{-7}$ & $2.49\times10^{-4}$ & $2.15\times10^{-4}$ \\
WSP & $1.10\times10^{-5}$ & $6.02\times10^{-7}$ & $7.78\times10^{-6}$ & $7.10\times10^{-6}$ \\
WSP-PC & $4.63\times10^{-5}$ & $7.51\times10^{-7}$ & $3.27\times10^{-5}$ & $4.79\times10^{-5}$ \\
PtW & $\mathbf{4.31\times10^{-17}}$ & $\mathbf{3.14\times10^{-33}}$ & $\mathbf{0}$ & $\mathbf{5.73\times10^{-17}}$ \\
\bottomrule
\end{tabular}
\end{table}

\begin{table}
\caption{Quantitative results on the advection equation. Values are medians over three runs.}
\label{tab:advection}
\centering
\scriptsize
\setlength{\tabcolsep}{3pt}
\begin{tabular}{lcccc}
\toprule
Method & Rel. L2 $\downarrow$ & PDE Res. $\downarrow$ & IC $\downarrow$ & BC $\downarrow$ \\
\midrule
PINN & $2.86\times10^{-1}$ & $7.37\times10^{-5}$ & $1.76\times10^{-2}$ & $8.32\times10^{-2}$ \\
WSP & $3.88\times10^{-4}$ & $2.54\times10^{-29}$ & $2.72\times10^{-4}$ & $2.15\times10^{-5}$ \\
WSP-PC & $3.88\times10^{-4}$ & $2.55\times10^{-29}$ & $2.72\times10^{-4}$ & $2.15\times10^{-5}$ \\
PtW & $\mathbf{2.00\times10^{-17}}$ & $\mathbf{0}$ & $\mathbf{0}$ & $\mathbf{1.75\times10^{-16}}$ \\
\bottomrule
\end{tabular}
\end{table}

\subsection{Heat Equation}

% P1. 定量評価：PtWの解の近似精度と物理整合性を比較し、アーキテクチャ編集による解析的構造の導入が高精度化に寄与することを示す。
Table~\ref{tab:heat} presents the quantitative results on the heat equation.
PtW achieves a relative $L_2$ error of $4.31\times10^{-17}$, substantially outperforming all baseline methods.
The PDE residual and initial and boundary condition errors are also nearly zero.
This high accuracy is primarily attributed to architecture editing, which incorporates the analytical structure of the heat equation into the NN representation.

% P2. 視覚評価：予測解と参照解の一致度および誤差分布を比較し、PtWが機械精度に近い解を再現することを示す。
Figure~\ref{fig:solutions} compares the predicted spatiotemporal solutions and pointwise absolute errors.
While all methods visually reproduce the reference solution, the baseline methods exhibit nonzero errors across the spatiotemporal domain.
In contrast, PtW reproduces the reference solution with pointwise errors approaching machine precision, consistent with the quantitative results.

\subsection{Advection Equation}

% P1. 定量評価：PtWは機械精度に近い解を達成し、アーキテクチャ編集による特性座標の導入が高精度化に寄与する。
Table~\ref{tab:advection} presents the quantitative results on the advection equation.
PtW achieves a relative $L_2$ error of $2.00\times10^{-17}$, substantially outperforming all baseline methods.
The PDE residual and initial and boundary condition errors are also nearly zero.
This high accuracy is primarily attributed to architecture editing, which incorporates the characteristic coordinate transformation into the NN representation.

\begin{table}
\caption{Quantitative results on Burgers' equation. Values are medians over three runs. PDE residual MSE is evaluated on a $256\times101$ grid and is sensitive to spatial resolution for PtW.}
\label{tab:burgers}
\centering
\scriptsize
\setlength{\tabcolsep}{3pt}
\begin{tabular}{lcccc}
\toprule
Method & Rel. L2 $\downarrow$ & PDE Res. $\downarrow$ & IC $\downarrow$ & BC $\downarrow$ \\
\midrule
PINN & $4.69\times10^{-2}$ & $\mathbf{4.44\times10^{-3}}$ & $1.29\times10^{-2}$ & $9.44\times10^{-4}$ \\
WSP & $2.21\times10^{-2}$ & $3.66\times10^{-1}$ & $9.96\times10^{-3}$ & $2.45\times10^{-4}$ \\
WSP-PC & $7.17\times10^{-2}$ & $7.07\times10^{-1}$ & $1.41\times10^{-3}$ & $2.71\times10^{-4}$ \\
PtW & $\mathbf{1.53\times10^{-2}}$ & $1.43\times10^{1}$ & $\mathbf{4.90\times10^{-15}}$ & $\mathbf{6.39\times10^{-7}}$ \\
\bottomrule
\end{tabular}
\end{table}

\begin{table}
\caption{Quantitative results on the Allen--Cahn equation. Values are medians over three runs.}
\label{tab:allen_cahn}
\centering
\scriptsize
\setlength{\tabcolsep}{3pt}
\begin{tabular}{lcccc}
\toprule
Method & Rel. L2 $\downarrow$ & PDE Res. $\downarrow$ & IC $\downarrow$ & BC $\downarrow$ \\
\midrule
PINN & $5.20\times10^{-1}$ & $\mathbf{1.44\times10^{-3}}$ & $5.87\times10^{-2}$ & $2.53\times10^{-3}$ \\
WSP & $7.45\times10^{-2}$ & $1.44\times10^{-1}$ & $3.30\times10^{-2}$ & $1.71\times10^{-2}$ \\
WSP-PC & $2.29\times10^{-1}$ & $2.86\times10^{-1}$ & $1.85\times10^{-2}$ & $7.63\times10^{-5}$ \\
PtW & $\mathbf{3.53\times10^{-2}}$ & $1.67\times10^{-2}$ & $\mathbf{8.13\times10^{-17}}$ & $\mathbf{4.88\times10^{-16}}$ \\
\bottomrule
\end{tabular}
\end{table}

% P2. 視覚評価：移流する波形の再現性と誤差分布を比較し、PtWが機械精度に近い解を再現することを示す。
Figure~\ref{fig:solutions} compares the predicted spatiotemporal solutions and pointwise absolute errors.
While PINN exhibits noticeable deviations from the reference solution, WSP and WSP-PC reproduce the traveling wave with smaller errors.
PtW further reduces these errors to machine precision, accurately preserving the wave propagation throughout the spatiotemporal domain.

\begin{figure*}
\centering
\includegraphics[width=0.9\textwidth]{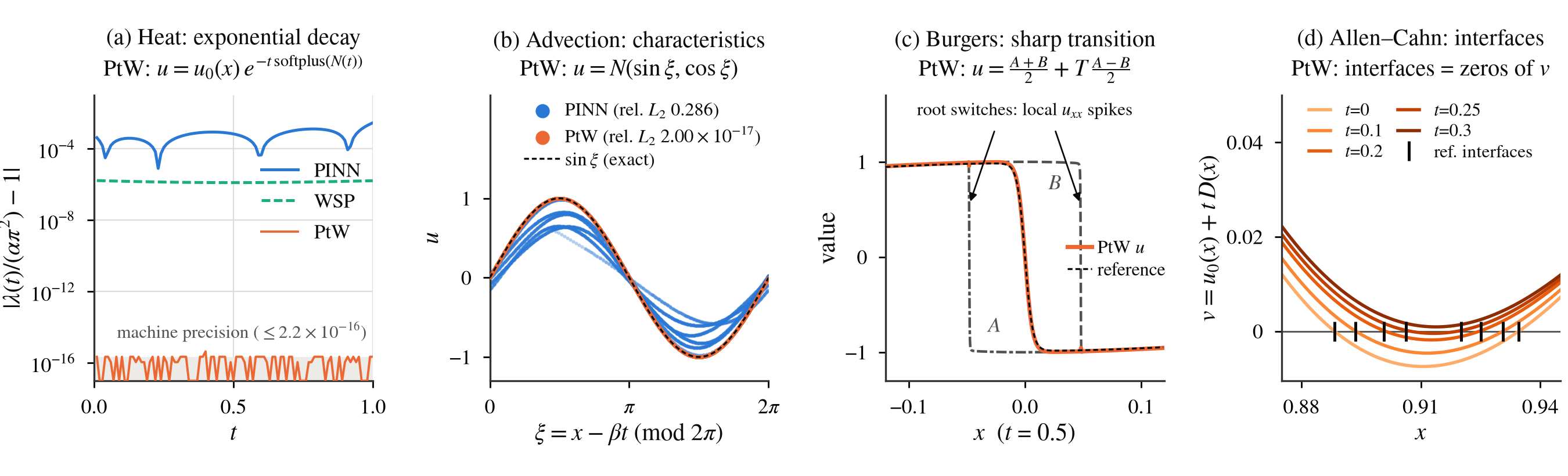}
\vspace{-8mm}
\caption{Structures inside the networks generated by PtW (median runs).
(a) Heat: relative error of the decay rate $\lambda(t)$ of the fundamental mode; the network computes
$u=u_0(x)\,e^{-t\,\mathrm{softplus}(N(t))}$.
(b) Advection: $u$ against the characteristic coordinate $\xi=x-\beta t$; the network computes
$u=N(\sin\xi,\cos\xi)$ and collapses onto $\sin\xi$, unlike PINN.
(c) Burgers' equation at $t=0.5$: two chains of MLPs, $A$ and $B$, iterate the characteristic
relation for the two sides of the shock, and a $\tanh$ switch $T$ combines them into
$u=\frac{A+B}{2}+T\,\frac{A-B}{2}$; where a chain switches roots, $u_{xx}$ has local spikes.
(d) Allen--Cahn: the network computes $u=\tanh(z)$ from $v=u_0(x)+t\,D(x)$ with a sine-activated
MLP $D$, where z is a monotone function of v, so the interfaces are the zeros of $v$; colored curves show $v$ at $t=0$--$0.3$ and black
ticks the reference interfaces.}
\label{fig:analysis}
\end{figure*}

\begin{table*}
\caption{Network size and computational cost (medians over three runs). Time is the wall-clock time
per network, including PINN training (PINN, PINN-WR) and all LLM calls; LLM times depend on server
load. For WSP-PC on Burgers' equation, two of the three runs include a first call that exceeded
the output-token limit after 82 and 92 minutes before a successful retry.}
\label{tab:cost}
\centering
\scriptsize
\setlength{\tabcolsep}{3pt}
\begin{tabular}{lccccccccc}
\toprule
& & \multicolumn{2}{c}{Heat} & \multicolumn{2}{c}{Advection} & \multicolumn{2}{c}{Burgers'} & \multicolumn{2}{c}{Allen--Cahn} \\
\cmidrule(lr){3-4}\cmidrule(lr){5-6}\cmidrule(lr){7-8}\cmidrule(lr){9-10}
Method & LLM calls & params & time [s] & params & time [s] & params & time [s] & params & time [s] \\
\midrule
PINN & 0 & 501 & 186 & 501 & 136 & 501 & 187 & 501 & 247 \\
%PINN-WR & 5 & 501 & 234 & 501 & 460 & 501 & 321 & 501 & 341 \\
%\midrule
WSP & 1 & 501 & 290 & 501 & 121 & 501 & 1009 & 501 & 705 \\
WSP-PC & 1 & 501 & 269 & 501 & 144 & 501 & 5909 & 501 & 2037 \\
%WSP-WR & 6 & 501 & 997 & 501 & 357 & 501 & 1272 & 501 & 873 \\
%WSP-PC-WR & 6 & 501 & 901 & 501 & 418 & 501 & 6171 & 501 & 2268 \\
%WSP-PC-AE & 2--3 & 4--33 & 31 & 4--17 & 33 & 104--342 & 502 & 36--52 & 427 \\
PtW & 7--8 & 4--33 & 76 & 4--17 & 70 & 104--342 & 1066 & 36--52 & 588 \\
\bottomrule
\end{tabular}
\end{table*}

\subsection{Burgers' Equation}

% P1. 定量評価：PtWは高い解の近似精度を達成する一方、PDE残差は局所的な大きな値に支配され、評価格子に強く依存する。
Table~\ref{tab:burgers} presents the quantitative results on Burgers' equation.
PtW achieves the lowest relative $L_2$ error of $1.53\times10^{-2}$ among the methods in Table~\ref{tab:burgers}.
It also achieves the lowest initial and boundary condition errors.
However, its PDE residual MSE is substantially higher than that of PINN.
This discrepancy is caused by a few localized residual peaks: the median squared residual of PtW is lower than that of PINN ($4.4\times10^{-6}$ versus $1.9\times10^{-5}$), whereas its MSE increases by orders of magnitude on a finer spatial grid.
For one PtW run, increasing the spatial resolution from 256 to 2001 points raises the MSE from $5.45\times10^{-2}$ to $6.6\times10^{9}$, while the PINN MSE changes by less than 20\%.

% P2. 視覚評価：PtWは非線形な解の構造を良好に再現するが、急峻な変化が生じる領域に誤差が集中する。
Figure~\ref{fig:solutions} compares the predicted solutions and pointwise absolute errors.
The errors of PINN concentrate along the shock at $x=0$ and persist after the shock has formed,
whereas those of WSP and WSP-PC spread over the smooth region.
PtW has the lowest error away from the shock (root-mean-square error $4.8\times10^{-3}$ for
$|x|\ge0.05$, versus $8.0\times10^{-3}$ for PINN) and at the shock after it has formed
($1.3\times10^{-2}$ for $|x|<0.05$ and $t\ge0.4$, versus $1.6\times10^{-1}$ for PINN).
Its error concentrates near $x=0$ while the shock forms, where the maximum pointwise error of
$0.19$ occurs at $t=0.33$.

\subsection{Allen--Cahn Equation}

% P1. 定量評価：PtWはPINN・WSPより高い解の近似精度と初期・境界条件の整合性を達成する一方、PDE残差ではPINNが優れる。
Table~\ref{tab:allen_cahn} presents the quantitative results on the Allen--Cahn equation.
PtW achieves a relative $L_2$ error of $3.53\times10^{-2}$, outperforming PINN, WSP, and WSP-PC.
It also achieves near-zero initial and boundary condition errors.
However, PINN achieves a lower PDE residual than PtW, indicating that improved solution accuracy does not necessarily correspond to better satisfaction of the governing equation.

% P2. 視覚評価：PtWとWSP-PCはいずれも界面消滅を再現するが、PtWは解全体の近似精度とエネルギー誤差で優れる。
Figure~\ref{fig:allen_cahn} compares the predicted spatiotemporal solutions.
Both PtW and WSP-PC reproduce the interface annihilation observed in the reference solution.
However, PtW achieves lower solution and energy errors than WSP-PC, indicating a more accurate approximation of the overall interface dynamics.

\begin{table*}
\caption{Ablation study of physics-constrained prompting (PC), architecture editing (AE), and weight refinement (WR). Values are medians over three runs. The best result for each metric and PDE is shown in bold.}
\label{tab:ablation}
\centering
\scriptsize
\setlength{\tabcolsep}{3pt}
\resizebox{\textwidth}{!}{
\begin{tabular}{lcccccccccccccccc}
\toprule
& \multicolumn{4}{c}{Heat} & \multicolumn{4}{c}{Advection} & \multicolumn{4}{c}{Burgers'} & \multicolumn{4}{c}{Allen--Cahn} \\
\cmidrule(lr){2-5}\cmidrule(lr){6-9}\cmidrule(lr){10-13}\cmidrule(lr){14-17}
Method & L2 & PDE & IC & BC & L2 & PDE & IC & BC & L2 & PDE & IC & BC & L2 & PDE & IC & BC \\
\midrule
PINN
& 2.28e-4 & 6.45e-7 & 2.49e-4 & 2.15e-4
& 2.86e-1 & 7.37e-5 & 1.76e-2 & 8.32e-2
& 4.69e-2 & \textbf{4.44e-3} & 1.29e-2 & 9.44e-4
& 5.20e-1 & \textbf{1.44e-3} & 5.87e-2 & 2.53e-3 \\
PINN-WR
& 2.28e-4 & 6.45e-7 & 2.49e-4 & 2.15e-4
& 2.38e-7 & 2.45e-29 & 1.64e-7 & 1.60e-8
& 4.69e-2 & \textbf{4.44e-3} & 1.29e-2 & 9.44e-4
& 5.20e-1 & \textbf{1.44e-3} & 5.87e-2 & 2.53e-3 \\
\midrule
WSP
& 1.10e-5 & 6.02e-7 & 7.78e-6 & 7.10e-6
& 3.88e-4 & 2.54e-29 & 2.72e-4 & 2.15e-5
& 2.21e-2 & 3.66e-1 & 9.96e-3 & 2.45e-4
& 7.45e-2 & 1.44e-1 & 3.30e-2 & 1.71e-2 \\
WSP-PC
& 4.63e-5 & 7.51e-7 & 3.27e-5 & 4.79e-5
& 3.88e-4 & 2.55e-29 & 2.72e-4 & 2.15e-5
& 7.17e-2 & 7.07e-1 & 1.41e-3 & 2.71e-4
& 2.29e-1 & 2.86e-1 & 1.85e-2 & 7.63e-5 \\
WSP-WR
& 2.55e-7 & 2.77e-13 & 1.19e-7 & 1.14e-7
& 1.14e-9 & 2.44e-29 & 5.18e-10 & 2.63e-9
& 2.17e-2 & 1.60e-1 & 9.94e-3 & 2.45e-4
& 1.29e-1 & 3.24e-2 & 4.11e-2 & 2.63e-4 \\
WSP-PC-WR
& 4.58e-7 & 4.99e-13 & 1.92e-7 & 2.18e-7
& 1.61e-7 & 2.48e-29 & 1.14e-7 & 3.19e-9
& 9.62e-2 & 4.68e-1 & 1.41e-3 & 2.65e-4
& 1.39e-1 & 7.47e-2 & 3.29e-2 & 1.00e-4 \\
WSP-PC-AE
& 1.21e-14 & 1.70e-28 & \textbf{0} & \textbf{5.73e-17}
& 2.64e-9 & \textbf{0} & 1.86e-9 & 1.86e-16
& \textbf{1.17e-2} & 9.24e+1 & \textbf{1.65e-16} & \textbf{3.00e-12}
& \textbf{3.49e-2} & 3.31e-2 & \textbf{8.13e-17} & 5.03e-16 \\
PtW
& \textbf{4.31e-17} & \textbf{3.14e-33} & \textbf{0} & \textbf{5.73e-17}
& \textbf{2.00e-17} & \textbf{0} & \textbf{0} & \textbf{1.75e-16}
& 1.53e-2 & 1.43e+1 & 4.90e-15 & 6.39e-7
& 3.53e-2 & 1.67e-2 & \textbf{8.13e-17} & \textbf{4.88e-16} \\
\bottomrule
\end{tabular}
}
\end{table*}

\begin{table}
\caption{Physical constraint violations. $V$: median number of violated constraints among C1--C5
(Appendix~\ref{app:constraints}) over three runs, with the number of runs with $V=0$ in
parentheses. Phys.: number of runs whose problem-specific physical metric
(Table~\ref{tab:physics}) is within tolerance.}
\label{tab:violations}
\centering
\scriptsize
\setlength{\tabcolsep}{3pt}
\begin{tabular}{lcccccccc}
\toprule
& \multicolumn{2}{c}{Heat} & \multicolumn{2}{c}{Advection} & \multicolumn{2}{c}{Burgers'} & \multicolumn{2}{c}{Allen--Cahn} \\
\cmidrule(lr){2-3}\cmidrule(lr){4-5}\cmidrule(lr){6-7}\cmidrule(lr){8-9}
Method & $V$ & Phys. & $V$ & Phys. & $V$ & Phys. & $V$ & Phys. \\
\midrule
PINN & 0 (3/3) & 3/3 & 4 (0/3) & 0/3 & 3 (0/3) & 3/3 & 3 (0/3) & 1/3 \\
PINN-WR & 0 (3/3) & 3/3 & 0 (3/3) & 3/3 & 3 (0/3) & 3/3 & 3 (0/3) & 1/3 \\
\midrule
WSP & 0 (3/3) & 3/3 & 0 (3/3) & 3/3 & 3 (0/3) & 3/3 & 2 (0/3) & 0/3 \\
WSP-PC & 0 (3/3) & 3/3 & 0 (3/3) & 3/3 & 1 (0/3) & 1/3 & 0 (3/3) & 3/3 \\
WSP-WR & 0 (3/3) & 3/3 & 0 (3/3) & 3/3 & 3 (0/3) & 3/3 & 1 (0/3) & 0/3 \\
WSP-PC-WR & 0 (3/3) & 3/3 & 0 (3/3) & 3/3 & 0 (3/3) & 1/3 & 0 (3/3) & 3/3 \\
WSP-PC-AE & 0 (3/3) & 3/3 & 0 (3/3) & 3/3 & 0 (2/3) & 3/3 & 1 (1/3) & 3/3 \\
PtW & 0 (3/3) & 3/3 & 0 (3/3) & 3/3 & 0 (3/3) & 2/3 & 0 (3/3) & 3/3 \\
\bottomrule
\end{tabular}
\end{table}

\subsection{Analysis of Generated Weights and Architectures}\label{sec:analysis}

% P1. 全体分析：LLMが生成したNNの構造とパラメータ数を比較し、アーキテクチャ編集の特徴を明らかにする。
To investigate how PtW represents PDE solutions, we analyze the generated NN architectures and weights, as shown in Fig.~\ref{fig:analysis} and Table~\ref{tab:cost}.
The edited networks contain substantially fewer parameters than the initial architecture with 501 parameters.
Specifically, the generated networks have 4--33 parameters for the heat equation, 4--17 for advection, 104--342 for Burgers', and 36--52 for Allen--Cahn.

% P2. 線形PDE：HeatとAdvectionでは解析的な解の構造をアーキテクチャに組み込むことで高い精度を達成する。
For the heat and advection equations, the LLM incorporates analytical solution structures into the NN architectures.
For the heat equation, the generated network represents exponential decay using $u=u_0(x)\exp(-t\,\mathrm{softplus}(N(t)))$, where the MLP $N$ produces a constant decay rate corresponding to $\alpha\pi^2$ (Fig.~\ref{fig:analysis}a).
For advection, the network uses the characteristic coordinate $\xi=x-\beta t$ and represents the traveling wave through an MLP applied to $(\sin\xi,\cos\xi)$ (Fig.~\ref{fig:analysis}b).
These analytical structures contribute substantially to the near-machine-precision solution accuracy.

% P3. 非線形PDE：Burgers'とAllen--Cahnでは、反復計算や非線形MLPによって複雑な解の構造を表現する。
For Burgers' equation, the generated networks approximate the characteristic relation $u=u_0(x-ut)$ through unrolled iterative computations using multiple small MLPs.
The median network combines two iterative branches with a $\tanh$-based transition function to represent the nonlinear solution structure (Fig.~\ref{fig:analysis}c).
However, abrupt root switches within the unrolled iterations produce isolated points with very large
second derivatives (Fig.~\ref{fig:analysis}c), which dominate the PDE residual despite the low
solution error.
For Allen--Cahn, the generated network represents the evolving spatial field using a nonlinear MLP and applies a $\tanh$ transformation to model the interfaces (Fig.~\ref{fig:analysis}d).
The nonlinear MLP structure is important for representing the interface dynamics, as replacing it with affine transformations increases the solution error.

% P4. 考察：解析的構造を組み込める線形PDEでは高精度化が顕著だが、非線形PDEでは局所的な物理整合性に課題が残る。
These results suggest that physics-guided architecture editing is particularly effective when analytical solution structures can be incorporated into the NN representation.
For nonlinear PDEs, the generated architectures can achieve accurate global solution approximations, although local violations of the governing equations remain a challenge.

\subsection{Ablation Study}\label{sec:ablation}

% P1. PCの効果：PC単独では解の近似精度は改善しないが、物理制約違反を減少させる。
Table~\ref{tab:ablation} presents the ablation results for PC, AE, and WR.
We first evaluate the effect of PC by comparing WSP and WSP-PC.
PC does not improve the relative $L_2$ error across the four PDEs.
However, it reduces the median number of violated physical constraints from 3 to 1 for Burgers' equation and from 2 to 0 for the Allen--Cahn equation (Table~\ref{tab:violations}).
These results indicate that PC improves physical constraint satisfaction rather than solution accuracy.

% P2. AEの効果：物理構造をNNに組み込むAEは、特に線形PDEにおいて解の近似精度を大幅に改善する。
We next investigate the effect of AE, which is combined with PC in our experiments.
Compared with WSP-PC, WSP-PC-AE substantially reduces the relative $L_2$ error across all four PDEs.
The improvements are particularly significant for the heat and advection equations, where analytical solution structures can be incorporated into the NN architectures.
For Burgers' and Allen--Cahn equations, AE also improves solution accuracy, demonstrating its effectiveness for nonlinear PDEs.

% P3. WRの効果：WSPへの適用ではPDEごとに効果が異なり、PINN-WRではAdvectionのみ解析的構成への書き換えによって大幅に改善する。
Finally, we evaluate the effect of WR.
When applied to WSP, WR improves solution accuracy for the heat and advection equations, while providing limited or negative improvements for Burgers' and Allen--Cahn equations.
When applied to trained PINN weights, WR improves only the advection equation, reducing the relative $L_2$ error from $2.86\times10^{-1}$ to $2.38\times10^{-7}$.
In all three runs, the LLM discarded the PINN weights in the first refinement round and replaced them with an analytical construction.
For the other three PDEs, PINN-WR produces the same results as PINN.
When combined with PC and AE, WR increases the number of runs satisfying all physical constraints from 2/3 to 3/3 for Burgers' equation and from 1/3 to 3/3 for the Allen--Cahn equation.
However, these improvements do not necessarily reduce the relative $L_2$ error, as the final model is selected by prioritizing physical constraint satisfaction.

% P4. まとめ：AEが解の精度向上の主要因であり、PCとWRは主に物理制約の充足に寄与する。
Overall, AE provides the largest improvements in solution accuracy, particularly for PDEs with known analytical structures.
PC mainly improves constraint satisfaction, while WR improves accuracy for heat and advection and, combined with PC, constraint satisfaction.

\subsection{Limitations}\label{sec:limitations}
For the heat and advection equations, PtW achieves near-machine-precision accuracy by incorporating known analytical solution structures into the NN architectures.
Therefore, these results do not necessarily demonstrate the ability to generate highly accurate weights for PDEs without known analytical solutions.
For Burgers' equation, the generated networks exhibit large local derivatives, resulting in high PDE residuals despite relatively accurate solution approximations.
These residuals are highly sensitive to spatial grid resolution and should not be interpreted as grid-converged estimates.
Furthermore, our evaluation is limited to one-dimensional PDEs, a single LLM, and three runs per method.
Future work will investigate more complex PDEs without known analytical solutions, improve the physical consistency of generated networks, and explore the integration of LLM-generated architectures with gradient-based optimization.

\section{Conclusion}\label{sec:conclusion}
We proposed PtW, a framework for solving PDEs by directly generating NN architectures and weights using an LLM without gradient-based optimization.
Experiments on four PDEs demonstrated that PtW achieves lower solution errors than conventional PINNs and WSP, while satisfying the specified physical constraints.
Ablation studies showed that architecture editing is the primary contributor to solution accuracy, whereas physics-constrained prompting and weight refinement mainly improve constraint satisfaction.
However, relatively large PDE residuals remain for nonlinear PDEs, indicating limitations in physical consistency.

\bibliography{example_paper}
\bibliographystyle{icml2025}

\newpage
\appendix
\onecolumn

\section{PDE Settings}\label{app:pde_settings}

All problems are one-dimensional in space with $t\in[0,1]$.
The physical constraints used for each problem are summarized in Table~\ref{tab:constraints}.

\subsection{Heat Equation}
We consider $u_t=\alpha u_{xx}$ with $\alpha=0.1$ on $x\in[0,1]$, subject to $u(x,0)=\sin(\pi x)$ and $u(0,t)=u(1,t)=0$.
The reference solution is given by $u(x,t)=e^{-\alpha\pi^2t}\sin(\pi x)$.

\subsection{Advection Equation}
We consider $u_t+\beta u_x=0$ with $\beta=30$ on $x\in[0,2\pi]$, subject to $u(x,0)=\sin x$ and periodic boundary conditions.
The reference solution is given by $u(x,t)=\sin(x-\beta t)$.

\subsection{Burgers' Equation}
We consider $u_t+uu_x=\nu u_{xx}$ with $\nu=0.01/\pi$ on $x\in[-1,1]$, subject to $u(x,0)=-\sin(\pi x)$ and $u(\pm1,t)=0$.
The solution develops a steep viscous shock near $x=0$.
The reference solution is computed using the Cole--Hopf transformation with numerical quadrature.

\subsection{Allen--Cahn Equation}
We consider $u_t=d\,u_{xx}+\kappa(u-u^3)$ with $d=10^{-3}$ and $\kappa=5$ on $x\in[-1,1]$, subject to periodic boundary conditions for $u$ and $u_x$.
The initial condition is defined as
\begin{equation}
u(x,0)=\sum_{k=1}^{5}A_k\sin(k\pi x+\phi_k),
\end{equation}
where $A=(-0.3337,-0.1499,0.1401,0.0331,-0.1462)$ and $\phi=(2.7214,3.0100,1.0037,4.6155,0.7142)$.
The initial field contains six interfaces, which evolve and undergo annihilation during the simulation.
The reference solution is computed using a Fourier pseudo-spectral method with 2048 spatial grid points and fourth-order exponential time differencing (ETDRK4)~\cite{kassam2005fourth} with a time step of $5\times10^{-4}$.

\section{Implementation Details}\label{app:implementation}

\paragraph{LLM.}
All LLM-based methods use Claude Opus 5.5 (\texttt{claude-opus-5-5}) with high reasoning effort.
The model is accessed through Claude Code in non-interactive mode without tool access.
Architecture editing, weight generation, and iterative refinement are performed within a single LLM session.
The LLM generates architectures and weights without executing code or performing numerical optimization.
Each run consists of a single generation step (preceded by architecture editing for AE) and, for
methods with WR, $K=5$ refinement rounds; each LLM call has a time limit of one or two hours.

\paragraph{Prompts.}
All LLM-based methods receive the governing equation, computational domain, initial and boundary conditions, NN architecture, and evaluation criteria.
PC additionally specifies physical constraints, while AE provides architecture editing rules and supported operations.
The LLM outputs the generated weights in JSON format.
Reference solutions and solution errors are not provided to the LLM.
%The complete prompts are included in the supplementary material.

\paragraph{Initial architecture.}
The initial architecture $\mathcal{A}_0$ is a fully connected NN with two hidden layers of 20 $\tanh$ units:
\begin{equation}
u_{\boldsymbol{\theta},\mathcal{A}_0}(x,t)
=W_3\tanh\left(W_2\tanh\left(W_1[x,t]^\top+b_1\right)+b_2\right)+b_3.
\end{equation}
The network contains 501 parameters and is evaluated in double precision without coordinate normalization.

\paragraph{Physics-informed loss and feedback.}
The physics-informed loss is defined as $\mathcal{L}=\mathcal{L}_{\mathrm{PDE}}+\mathcal{L}_{\mathrm{IC}}+\mathcal{L}_{\mathrm{BC}}$, using 4096 PDE collocation points, 256 initial points, and 128 boundary times.
All sampling points are fixed within each run, and derivatives are computed using automatic differentiation.
For periodic boundary conditions, the loss penalizes mismatches between the two boundaries, including first-order derivatives for Allen--Cahn.
During WR, the LLM receives the loss components, PDE residual diagnostics, initial and boundary condition errors, and coarse-grid predictions.
When PC is enabled, physical constraint evaluations are also included in the feedback.

\paragraph{Baselines.}
PINN is trained using Adam with a learning rate of $10^{-3}$ for 10,000 steps, followed by L-BFGS with a maximum of 10,000 iterations.
The PINN baseline uses the same initial architecture and collocation points as the LLM-based methods.
PINN-WR applies iterative weight refinement to pretrained PINN weights without gradient-based optimization during refinement.

\paragraph{Evaluation.}
All methods are evaluated on a uniform $256\times101$ spatiotemporal grid.
Solution accuracy is measured using the relative $L_2$ error.
Physical consistency is evaluated using the mean squared PDE residual and the root-mean-square initial and boundary condition errors.
Each method is evaluated over three independent runs, and the median of each metric is reported.
The visualizations correspond to the run with the median relative $L_2$ error.

\section{Physical Constraints}\label{app:constraints}

% P1. 物理制約：各PDEに対して、初期・境界条件、保存則、対称性などの物理制約を定義する。
Physics-constrained prompting (PC) introduces five physical constraints for each PDE, as summarized in Table~\ref{tab:constraints}.
These constraints are derived from the governing equations and their initial and boundary conditions.
Each constraint is evaluated independently from the generated NN without using reference solutions.
A constraint is regarded as satisfied when its evaluation falls within the specified tolerance.

\begin{table*}[t]
\caption{Physical constraints used in PC. Each PDE is evaluated using five constraints (C1--C5) with specified tolerances.}
\label{tab:constraints}
\centering
\scriptsize
\setlength{\tabcolsep}{5pt}
\begin{tabular}{lp{0.85\textwidth}}
\toprule
PDE & Physical constraints \\
\midrule
Heat
& C1: Initial condition (maximum error $\le 0.01$).
C2: Dirichlet boundary conditions (maximum error $\le 0.01$).
C3: Maximum principle $0\le u\le 1$ (tolerance $0.01$).
C4: Non-increasing energy $E(t)=\int_0^1 u^2\,dx$ (increase $\le 10^{-3}E(0)$).
C5: Symmetry $u(x,t)=u(1-x,t)$ (maximum error $\le 0.01$). \\
\midrule
Advection
& C1: Initial condition (maximum error $\le 0.01$).
C2: Periodic boundary condition (maximum error $\le 0.01$).
C3: Maximum amplitude $|u|\le 1$ (tolerance $0.01$).
C4: Conservation of mass (absolute tolerance $0.04$) and $\int u^2\,dx$ (relative tolerance $1\%$).
C5: Transport speed of the first Fourier mode (relative tolerance $1\%$), with amplitude ratio $\ge 0.1$. \\
\midrule
Burgers'
& C1: Initial condition (maximum error $\le 0.01$).
C2: Dirichlet boundary conditions (maximum error $\le 0.01$).
C3: Maximum amplitude $|u|\le 1$ (tolerance $0.01$).
C4: Non-increasing energy $E(t)=\int_{-1}^{1}u^2\,dx$ (increase $\le 10^{-3}E(0)$).
C5: Odd symmetry $u(-x,t)=-u(x,t)$ (maximum error $\le 0.01$). \\
\midrule
Allen--Cahn
& C1: Initial condition, including interface positions (position tolerance $0.01$).
C2: Non-increasing number of interfaces.
C3: Maximum amplitude $|u|\le 1.01$.
C4: Non-increasing energy $E(t)=\int_{-1}^{1}[\frac{d}{2}u_x^2+\frac{\kappa}{4}(1-u^2)^2]\,dx$ (increase $\le 10^{-3}E(0)$).
C5: Interface slopes at $t=1$ within $[0.75,1.25]$ times the stationary kink slope. \\
\bottomrule
\end{tabular}
\end{table*}

% P2. 制約評価：制約違反数を定義し、WRにおけるフィードバックとモデル選択に利用する。
For each generated network, we define the number of violated constraints as
\begin{equation}
V(\boldsymbol{\theta})
=\sum_{i=1}^{5}
\mathbb{I}\left[C_i(\boldsymbol{\theta})\text{ is violated}\right],
\label{eq:constraint_violations}
\end{equation}
where $\mathbb{I}[\cdot]$ is the indicator function.
During WR, the evaluation results are provided to the LLM as feedback, and $V$ is used to prioritize physical constraint satisfaction in the final model selection.

\section{Architecture Editing Rules}\label{app:ae}

% P1. 編集形式：LLMは数式演算とMLPを組み合わせてNNのforward computationを定義する。
Physics-guided architecture editing (AE) allows the LLM to construct NN architectures by combining mathematical expressions and MLP blocks.
The architecture is represented as a sequence of named layers, followed by an output expression.
Each layer either evaluates a mathematical expression or applies an MLP to the outputs of preceding layers.

% P2. 許可する演算：物理方程式に基づく演算を許可しつつ、使用可能な関数や定数を制限する。
Expression layers support arithmetic operations, integer powers, and elementary functions, including $\sin$, $\cos$, $\exp$, $\log$, $\tanh$, and softplus.
Additional supported functions include $\sinh$, $\cosh$, sigmoid, $\sqrt{\cdot}$, erf, and artanh.
Expressions can use spatial and temporal coordinates, PDE coefficients, the initial condition $u_0(x)$, and outputs of preceding layers.
Numerical constants are restricted to integers between $-20$ and $20$, $0.5$, and $\pi$.
The coefficients of the initial condition cannot be directly introduced as numerical constants.

% P3. NN構造の制約：パラメータ数とネットワーク構造を制限し、編集後もNNとしての構造を保持する。
To preserve the NN representation, all trainable parameters must be weights or biases of MLP blocks.
Each MLP supports $\tanh$, $\sin$, sigmoid, or softplus activations, with at most four hidden layers and 64 units per hidden layer.
The output must depend on at least one MLP containing a hidden layer.
The total number of trainable parameters is limited to 501, matching the initial architecture $\mathcal{A}_0$.

% P4. 編集範囲と検証：数式の複雑さと微分可能性を制限し、生成されたアーキテクチャを検証する。
The architecture contains at most 40 layers, with at most 32 inputs per MLP.
Each MLP input expression is limited to 24 symbols, while the combined expression layers and output expression are limited to 40 symbols.
The output must be finite, twice differentiable in space, and once differentiable in time.
Non-differentiable operations such as comparisons, conditionals, absolute values, minimum, and maximum are not permitted.
The generated architecture is validated against these rules, and violations are returned to the LLM for correction, with at most two correction attempts.

\section{Additional Results}\label{app:results}

% P1. 追加評価：解の近似誤差に加えて、各PDEに固有の物理的挙動を定量的に評価する。
We further evaluate problem-specific physical properties of the generated solutions, as summarized in Table~\ref{tab:physics}.
For the heat equation, we measure the relative error of the decay rate of the fundamental mode.
For advection, we evaluate the propagation speed and amplitude of the first Fourier mode.
For Burgers' equation, we measure the shock formation time and the maximum spatial gradient.
For Allen--Cahn, we evaluate the interface annihilation time and the relative energy error.

\begin{table*}[t]
\caption{Problem-specific physical metrics. Values are medians over three runs, except for Allen--Cahn annihilation times, which are reported for individual runs. The reference row includes the corresponding tolerances.}
\label{tab:physics}
\centering
\scriptsize
\setlength{\tabcolsep}{3pt}
\resizebox{\textwidth}{!}{
\begin{tabular}{lccccccc}
\toprule
& Heat & \multicolumn{2}{c}{Advection} & \multicolumn{2}{c}{Burgers'} & \multicolumn{2}{c}{Allen--Cahn} \\
\cmidrule(lr){2-2}\cmidrule(lr){3-4}\cmidrule(lr){5-6}\cmidrule(lr){7-8}
Method & Decay err. & Speed err. & Amplitude err. & $t_{\mathrm{shock}}$ & $G(1)/G_{\mathrm{ref}}(1)$ & $t_{\mathrm{ann}}$ & Energy err. \\
\midrule
Reference (tol.)
& $0$ ($\le 0.01$) & $0$ ($\le 0.01$) & $0$ ($\le 0.05$)
& $0.23$ ($\pm 0.05$) & $1$ ($[0.5,2]$)
& $0.28$ ($\pm 0.05$) & $0$ \\
\midrule
PINN
& $5.80\times10^{-5}$ & $3.68\times10^{-3}$ & $0.445$
& $0.21$ & $0.99$ & $0.50$/$0.28$/-- & $0.216$ \\
WSP
& $6.89\times10^{-7}$ & $1.77\times10^{-7}$ & $2.21\times10^{-5}$
& $0.22$ & $1.13$ & --/--/-- & $0.0204$ \\
WSP-PC
& $1.96\times10^{-6}$ & $2.20\times10^{-7}$ & $3.28\times10^{-5}$
& $0.29$ & $1.05$ & $0.29$/$0.26$/$0.25$ & $0.133$ \\
PtW
& $2.22\times10^{-16}$ & $2.22\times10^{-16}$ & $4.44\times10^{-16}$
& $0.22$ & $1.01$ & $0.28$/$0.30$/$0.27$ & $8.78\times10^{-3}$ \\
\bottomrule
\end{tabular}
}
\end{table*}

% P2. 線形PDE：HeatとAdvectionでは、解析的構造の導入により時間発展の物理特性を高精度に再現する。
For the heat and advection equations, PtW achieves errors near machine precision in the decay rate, propagation speed, and amplitude.
These results are consistent with the analytical structures incorporated through architecture editing.

% P3. 非線形PDE：Burgers'とAllen--Cahnでは、衝撃波形成や界面消滅などの時間発展を評価する。
For Burgers' equation, PtW reproduces the shock formation time and final gradient magnitude close to the reference values.
For Allen--Cahn, PtW reproduces interface annihilation in all three runs, with annihilation times between $0.27$ and $0.30$, compared with $0.28$ for the reference.
PtW also achieves a lower relative energy error than the baseline methods.
These results demonstrate that the generated networks can reproduce important physical dynamics, although low PDE residuals are not guaranteed.

\end{document}